\documentclass[runningheads, orivec]{llncs}
\usepackage[T1]{fontenc}
\usepackage{graphicx}
\usepackage{booktabs}
\usepackage[misc]{ifsym}

\makeatletter
\newcommand{\samethanks}[1][\value{footnote}]{\footnotemark[#1]}
\makeatother
\usepackage{mwe}
\usepackage{amsmath}
\usepackage{amssymb}

\usepackage{microtype}
\usepackage{hyperref}

\usepackage{booktabs}
\usepackage{multirow}
\usepackage{tabularx}

\usepackage{graphicx}
\usepackage{subcaption}

\begin{document}

\title{Vision–Language-Guided Pseudo-Labels for Unsupervised Domain Adaptation in Semantic Segmentation for Waste Sorting}

\titlerunning{Vision–Language-Guided Pseudo-Labels for Domain Adaptation}

\author{
Udo Schlegel\inst{1,2}\thanks{Equal contribution. Contact via: \email{\{schlegel,shubhangi\}@dbs.ifi.lmu.de}}\Letter \and
Shubhangi\inst{1,2}\samethanks \and
Gabriel Dax\inst{3} \and
Sai Rahul Kaminwar\inst{3} \and \\
Florian Karl\inst{1,2,3} \and
Thomas Seidl\inst{1,2,3}
}

\authorrunning{Schlegel et al.}

\institute{LMU Munich, Munich, Germany,
\and
Munich Center for Machine Learning (MCML), Munich, Germany
\and
Fraunhofer Institute for Integrated Circuits IIS, Nuremberg, Germany}

\toctitle{Vision–Language-Guided Pseudo-Labels for Unsupervised Domain Adaptation in Semantic Segmentation for Waste Sorting}
\tocauthor{Udo Schlegel, Shubhangi, Gabriel Dax, Sai Rahul Kaminwar, Florian Karl, Thomas Seidl}

\maketitle              

\begin{abstract}
Obtaining labeled data for semantic segmentation in applied settings (e.g., autonomous driving, industrial waste sorting) is expensive and often infeasible at scale. 
We present a cross-modal pseudo-labeling pipeline that enables unsupervised domain adaptation without any target-domain annotations. 
The pipeline is built on two core foundation models: SAM generates class-agnostic region proposals, and EVA-CLIP assigns semantic labels based on region–text similarity, with confidence filtering ensuring that only reliable pseudo-labels are used for self-training a segmentation model. 
As an optional extension, BLIP provides language-grounded verification for ambiguous regions, thereby improving pseudo-label quality without altering the overall pipeline. 
Evaluated on two domain shifts, synthetic-to-real autonomous driving and, with a primary focus, lab-to-factory industrial waste sorting, the pipeline consistently improves over source-only baselines. 
Our results demonstrate that pseudo-label quality, not quantity, is a decisive factor in self-training under domain shift, and that cross-modal language grounding offers a practical path to reliable automatic annotation in deployment-critical applications.

\keywords{Vision-Language Models \and Pseudo-labeling \and Unsupervised Domain Adaptation \and Semantic Segmentation.}
\end{abstract}

\section{Introduction}
Pixel-wise semantic segmentation is central to applications such as autonomous driving and industrial inspection, yet dense annotations are expensive, especially under domain shift, where models must generalize from synthetic data to real scenes or from controlled lab imagery to factory environments. 
The main bottleneck is often not model training but the creation of pixel-wise labels for each new operating condition~\cite{csurka_unsupervised_2021}. 
This is particularly acute in industrial waste sorting, a lab-to-factory transfer setting where a model is trained on labeled lab images but deployed on conveyor-belt data whose appearance changes frequently. 
Continuously collecting dense annotations for new scenarios is costly because data annotation faces unique challenges in this setting, as conditions change with every run.
Compared with standard benchmarks, this scenario is dominated by complex backgrounds, occlusions, dirt, and visually similar fine-grained categories, which further increase annotation costs and render dense annotations impractical.

Unsupervised domain adaptation (UDA) provides a suitable framework for solving the underlying problem by using source and target domains.
Prior work on UDA for semantic segmentation has followed two main directions: 
adversarial alignment~\cite{tsai_learning_2020,vu_advent_2019}, which reduces the distributional gap by making target features or predictions indistinguishable from source ones, 
and self-training~\cite{zou_unsupervised_2018,zou_confidence_2019}, which iteratively generates pseudo-labels on unlabeled target images and retrains the segmenter on them. 
Recent transformer-based methods, such as DAFormer~\cite{hoyer_daformer_2022}, combine stronger backbones with carefully tuned pseudo-label filtering to push state-of-the-art performance on synthetic-to-real benchmarks like GTA5-to-Cityscapes~\cite{richter_playing_2016,cordts_cityscapes_2016}.
Despite these advances, all such methods share a fundamental limitation: they rely on the segmentation model's own confidence to supervise target-domain training~\cite{zhu_hard_2025}. 
Under a strong domain shift, this creates a circular dependency: the model must bootstrap reliable supervision from predictions that are themselves unreliable, a problem that confidence thresholding alone cannot resolve, particularly for fine-grained boundaries, rare object categories, and classes whose appearance changes across domains~\cite{zhao_unsupervised_2024}.
Experiments on unsupervised domain adaptation should be interpreted cautiously: validating DA methods without target labels is notoriously difficult and can yield overly optimistic performance estimates~\cite{ericsson_better_2023}. 
Ericsson et al.~\cite{ericsson_better_2023} further argue that many UDA methods are hard to deploy safely, because unresolved model-selection issues can lead to adapted models that perform worse than the source-only baseline.


We address these shortcomings by proposing a cross-modal pseudo-labeling pipeline that bypasses the segmentation model’s own confidence by using vision–language models as a source of target supervision.  
Instead of generating pseudo-labels from the segmenter itself, the pipeline chains pretrained components whose outputs are independent of the segmentation model. 
In industrial settings, UDA should operate end-to-end on unlabeled target streams, produce explicit pseudo-labels that can be inspected and versioned, and avoid heavyweight, backbone-specific training. 
Our goal is not to compete with heavily engineered UDA approaches, but to test whether a foundation-model chain can provide reliable supervision that improves a segmenter under domain shift, offering a robust, easy-to-deploy alternative for practitioners.
We thus use widely adopted baselines and treat the segmentation backbone as interchangeable, enabling the pseudo-label bank and verification stages to be plugged into stronger models without changing the overall pipeline. 
Concretely, the Segment Anything Model (SAM)~\cite{kirillov_segment_2023} generates high-quality class-agnostic region proposals, EVA-CLIP~\cite{sun_eva_2023} assigns semantic labels to each region via region–text similarity in a shared embedding space, and confidence filtering keeps only the most reliable region–label pairs for self-training, while an optional BLIP~\cite{li_blip_2022} captioning module provides language-grounded verification for low-confidence regions. 
The segmenter is then trained jointly on labeled source data and filtered pseudo-labeled target data, with all target supervision provided by this foundation-model chain.

\textbf{Contributions --}
We make the following contributions. 
First, we present an automatic cross-modal pseudo-labeling pipeline for UDA semantic segmentation that chains SAM~\cite{kirillov_segment_2023} for class-agnostic region proposals and EVA-CLIP~\cite{sun_eva_2023} for semantic label assignment via region–text similarity, requiring no target-domain annotations. 
Second, we show that pseudo-label quality is decisive for self-training under domain shift, and we demonstrate consistent improvements over a source-only DeepLabV3~\cite{chen_rethinking_2017} baseline on both a synthetic-to-real driving benchmark (GTA5-to-Cityscapes) and, with particular focus, a lab-to-factory industrial waste-sorting benchmark. 
Finally, we emphasize deployability: the pipeline produces an explicit pseudo-label bank for the target domain and is backbone-agnostic, enabling straightforward integration into existing segmentation systems\footnote{Repo:~~ \url{https://github.com/lmu-dbs/cross-modal-pseudo-labeling-pipeline}}.


\section{Related Work}
We discuss four central topics in the field of pseudo-labeling. 
First, unsupervised domain adaptation, especially for semantic segmentation, addresses shifts between source and target domains. 
Second, we review pseudo-labeling methods and the role of prediction confidence and confirmation bias in self-training under domain shift. 
Third, we introduce foundation models for the vision–language domain, which serve as the building blocks of our cross-modal pseudo-labeling pipeline. 
Fourth, we discuss open-vocabulary semantic segmentation methods that combine class-agnostic region proposals with VLMs and compare our closed-set, UDA-oriented pseudo-labeling approach against open-vocabulary ones.

\textbf{Unsupervised Domain Adaptation for Semantic Segmentation.}
Early UDA methods for semantic segmentation focused on aligning source and target distributions through adversarial training, either in the feature space~\cite{ganin_domain_2016} or in the output space~\cite{tsai_learning_2020}. 
Output-space adaptation proved particularly effective, as it aligns semantic predictions directly~\cite{tsai_learning_2020}. 
Additional regularization techniques, such as entropy minimization~\cite{vu_advent_2019} and self-attention distillation~\cite{li_content_2020,liu_structured_2019}, further improved metric performance and robustness.

Self-training with pseudo-labels has emerged as a powerful paradigm for unsupervised domain adaptation~\cite{zou_unsupervised_2018}. 
Class-Balanced Self-Training (CBST)~\cite{zou_confidence_2019} introduced confidence thresholding and class-balancing to mitigate confirmation bias. 
Recent transformer-based methods, such as DAFormer~\cite{hoyer_daformer_2022}, combine strong segmentation architectures with sophisticated pseudo-label filtering, and other pseudo-label refinement approaches~\cite{zhao_unsupervised_2024} further improve state-of-the-art results on synthetic-to-real benchmarks like GTA5-to-Cityscapes~\cite{richter_playing_2016,cordts_cityscapes_2016}. 
PLSR~\cite{zhao_unsupervised_2024} explicitly refines noisy pseudo-labels using an auxiliary refinement network. These methods mainly improve how pseudo-labels are used, filtered, or refined inside the segmentation training loop.
However, these approaches rely heavily on the segmentation model's own predictions and confidence scores, which can become unreliable under strong domain shift, especially for fine-grained boundaries, small objects, and rare class conditions characteristic of our lab-to-factory setting.

\textbf{Foundation Models for Vision and Vision--Language.}
SAM~\cite{kirillov_segment_2023} represents a breakthrough in zero-shot segmentation. Trained on the SA-1B dataset~\cite{kirillov_segment_2023}, SAM uses a promptable transformer architecture to generate object-like masks given sparse prompts (points, boxes) or in fully automatic mode. 
Its domain-agnostic priors make it uniquely suited for generating region proposals in shifted target domains where traditional segmenters fail.
Vision--language models (VLMs) bridge visual and textual understanding through contrastive pretraining on massive image--text pairs. 
CLIP~\cite{radford_learning_2021} learns a joint embedding space that enables zero-shot classification via text prompts. 
EVA-CLIP~\cite{sun_eva_2023} substantially improves zero-shot accuracy through advanced optimization, data augmentation, and scaling (e.g., EVA-CLIP achieves 82\% ImageNet zero-shot accuracy). 
Generative VLMs such as BLIP~\cite{li_blip_2022} and BLIP-2~\cite{li_blip_2023} extend this capability by generating free-form captions conditioned on images, providing richer semantic descriptions that can verify or refine contrastive predictions.

\textbf{Pseudo-Labeling Beyond Segmentation Confidence.}
Pseudo-labeling, treating model predictions on unlabeled data as ``pseudo ground truth'', is a cornerstone of semi-supervised and domain adaptation methods~\cite{lee_pseudo_2013}. 
While effective in moderate data regimes, its success critically depends on filtering noisy or overconfident predictions~\cite{xie_unsupervised_2020}. 
In domain adaptation, confirmation bias remains the primary failure mode: erroneous early pseudo-labels propagate and reinforce model errors~\cite{tarvainen_mean_2018,french_self_2018}. 
Recent work explores alternative supervision signals, including consistency regularization~\cite{french_semi_2020}, but these require architectural modifications or additional inductive biases. 
Our approach is unique in leveraging \emph{auxiliary foundation models} as an orthogonal source of pseudo-label supervision, bypassing the segmentation model's own (often unreliable) confidence.

\textbf{CLIP-based Pseudo-Labeling and Open-Vocabulary Semantic Segmentation.}
Recent work has increasingly explored vision--language models as sources of semantic supervision under limited annotation. 
In unsupervised domain adaptation, PADCLIP~\cite{lai_padclip_2023} leverages CLIP zero-shot predictions for pseudo-labeling and introduces adaptive debiasing. In contrast, our method does not adapt CLIP as the final task model, but uses CLIP-family models as external semantic verifiers for SAM-generated target regions. This produces an inspectable pseudo-label bank in which uncertain regions can be filtered or ignored before self-training, making our approach complementary to PADCLIP.

Our work is also closely related to open-vocabulary semantic segmentation (OVSS), which extends standard segmentation by allowing models to predict categories specified at test time through natural-language prompts, rather than restricting them to a fixed, closed label set~\cite{xu_simple_2022,zhu_survey_2024}. 
Recent OVSS methods typically pair a mask generator with a vision–language encoder such as CLIP, using text prompts or class names to classify class-agnostic regions, and often finetune CLIP on masked crops or introduce mask-aware adaptations to improve performance on unseen classes~\cite{liang_open_2022}. 
Conceptually, our cross-modal pseudo-labeling pipeline shares this two-stage structure (SAM for proposals and EVA-CLIP for region–text matching) but differs in goal and setting: we operate in a closed-set UDA regime and treat the vision–language predictions as external pseudo-labels for self-training a task-specific segmenter, rather than directly deploying the VLM as an open-vocabulary segmenter at inference time as seen in Zhu and Chen~\cite{zhu_survey_2024}. 
Although recent OVSS methods such as CAT-Seg~\cite{cho2024_catsegcostaggregationopenvocabulary} could in principle serve as stronger dense vision--language teachers, our current design keeps the VLM component upstream of the final segmenter.
This places our approach at the intersection of UDA and OVSS: we borrow the architectural insight that mask proposals plus CLIP region classification are powerful for semantic grounding, but we focus on how such open-vocabulary methods can act as a robust teacher to improve closed-vocabulary segmentation under severe industrial domain shift.

\begin{figure}[htb]
    \centering
    \includegraphics[width=.95\textwidth]{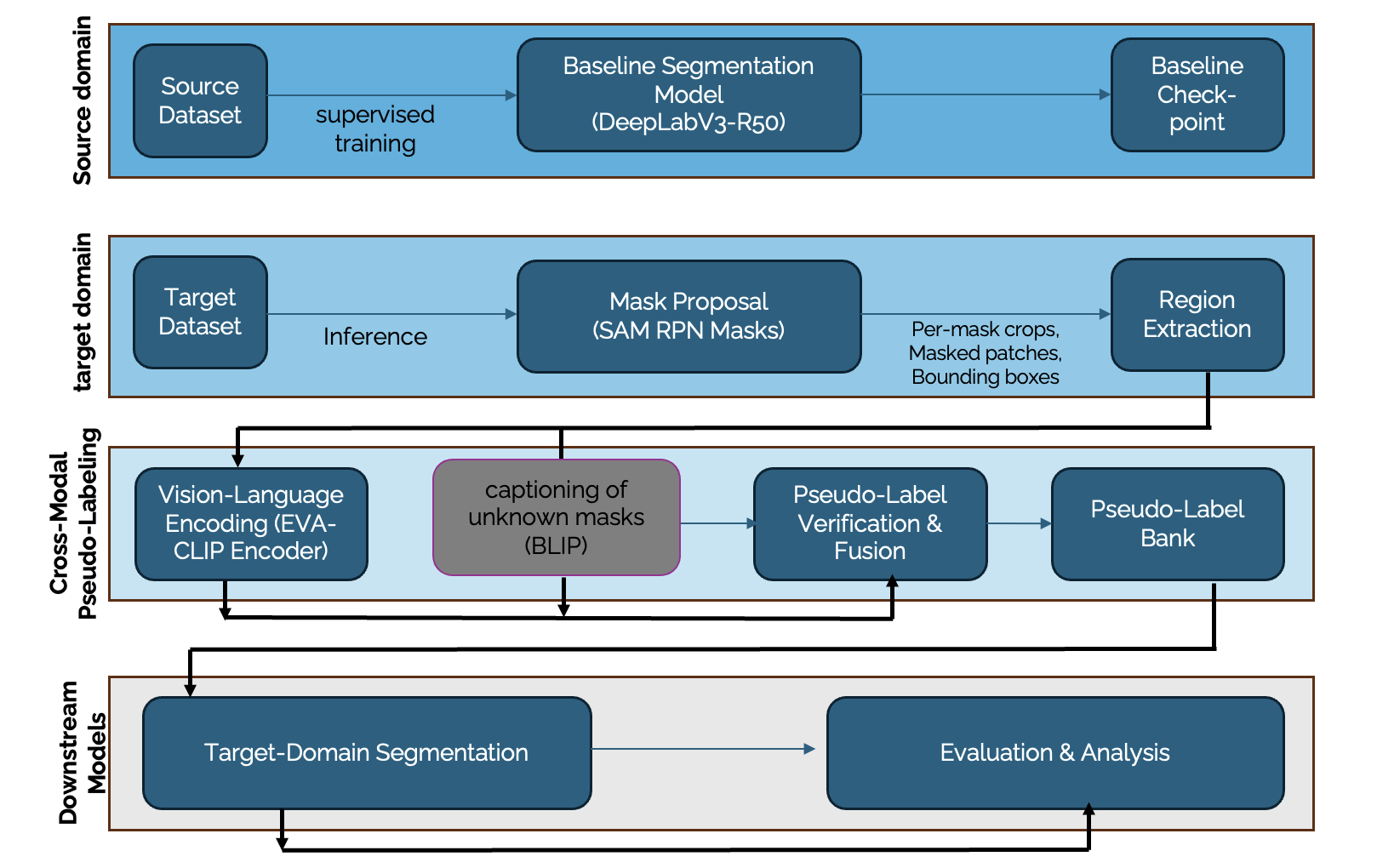}
    \caption{
    \textbf{Overview of the proposed cross-modal pseudo-labeling pipeline}, organized into four conceptual stages: 
    (i) supervised source-domain training of a baseline DeepLabV3 segmentation model, 
    (ii) target-domain mask proposal with SAM and region extraction from unlabeled target images, 
    (iii) cross-modal pseudo-label generation and verification using EVA-CLIP and optional BLIP captioning to populate a pseudo-label bank, and 
    (iv) downstream target-domain segmentation via self-training on the pseudo-labeled target data. 
    }
    \label{fig:pipeline}
\end{figure}

\section{Our proposed cross-modal pseudo-labeling pipeline}

\autoref{fig:pipeline} presents our proposed cross-modal pseudo-labeling pipeline, which is organized into four stages: 
\hyperref[sec:stage_one]{(i)} supervised source-domain training of a baseline segmentation model, 
\hyperref[sec:stage_two]{(ii)} target-domain mask proposal and region extraction, 
\hyperref[sec:stage_three]{(iii)} cross-modal pseudo-label generation and verification, and 
\hyperref[sec:stage_four]{(iv)} downstream target-domain self-training using the resulting pseudo-label bank. 
Together, these stages transform unlabeled target images into a structured supervision signal that can be used to adapt a standard segmentation network without any target-domain annotations. 
Because each stage is built from modular, off‑the‑shelf components (SAM~\cite{kirillov_segment_2023} for mask proposals, EVA-CLIP~\cite{sun_eva_2023} and optional BLIP~\cite{li_blip_2022} for semantic cues, and a conventional DeepLabV3~\cite{chen_rethinking_2017} segmenter), the pipeline is highly adaptable to new domains and can be implemented and adjusted by domain experts with limited prior experience in semantic segmentation architectures.
Moreover, each model block can be exchanged with alternative choices (e.g., different segmentation backbones, mask generators, or vision–language encoders) without changing the overall design, making the pipeline easy to repurpose for other application domains and computational budgets.

\subsection{Stage I: Source-Domain Baseline Segmentation}
\label{sec:stage_one}

In the first stage, we learn a baseline segmentation model on the labeled source domain. 
Concretely, we train DeepLabV3~\cite{chen_rethinking_2017} with a ResNet-50 backbone on
\[
D_S = \{(x_i^S, y_i^S)\}_{i=1}^{N_S},
\]
using a combination of pixel-wise cross-entropy loss and Dice loss. 
The cross-entropy term encourages correct class prediction at each pixel, while the Dice loss provides robustness to class imbalance while being less sensitive to tiny boundary errors due to optimization of region overlap~\cite{jadon_survey_2020}.

This training produces a “baseline checkpoint” that represents the best we can do when relying solely on source labels and no target information. 
We use this checkpoint in two ways: as the source-only reference in our quantitative comparisons and as the initialization for the self-training stage. 
This baseline segmenter does not generate pseudo-labels in our framework; supervision for the target domain comes instead from the cross-modal pseudo-labeling module.

\subsection{Stage II: Target-Domain Mask Proposal and Region Extraction}
\label{sec:stage_two}

The second stage operates purely on the unlabeled target dataset,
\[
D_T = \{x_j^T\}_{j=1}^{N_T},
\]
and aims to decompose each image into a set of candidate object or part regions. 
To this end, we use a frozen Segment Anything Model (SAM, ViT-H variant)~\cite{kirillov_segment_2023} in automatic mask generation mode. 
For each target image $x_j^T$, SAM returns a set of binary masks
\begin{equation}
    \{M_{j,k}\}_{k=1}^{K} = \text{SAM-Auto}(x_j^T), \quad M_{j,k} \in \{0,1\}^{H\times W},
\end{equation}
where $H$ and $W$ denote the image height and width, respectively.


To avoid overwhelming the later stages with noisy or redundant proposals, we apply lightweight post-processing: an \textit{area filter} that discards very small masks and \textit{non-maximum suppression}, which removes highly overlapping masks. 
This yields a diverse set of high-quality, object-like regions that provide good coverage of the scene. 
For each surviving mask $M_k$, we extract the region by cropping a context-aware bounding box around the mask, with a padding margin (adjustable via a parameter), and then apply the mask to the crop. 
The resulting region patches $x_k^T$ retain the local object while preserving some surrounding context, which is beneficial for semantic interpretation in the next stage.

\subsection{Stage III: Cross-Modal Pseudo-Labeling and Verification}
\label{sec:stage_three}

The third stage is the core cross-modal pseudo-labeling module, which converts SAM’s geometric proposals into semantic labels using a vision–language model.

\paragraph{EVA-CLIP semantic assignment --}
Each region patch $x_k^T$ is first encoded using the EVA-CLIP image encoder (ViT-L/336)~\cite{sun_eva_2023} to obtain a visual embedding
\begin{equation}
    v_k = \text{EVA-CLIP}_\text{img}(x_k^T) \in \mathbb{R}^{1024}.
\end{equation}
We define a fixed semantic label set shared between source and target domains, and for each class $c \in \{1,\ldots,C\}$ we construct a simple text prompt $t_c =$ “a photo of a \{class name\}”. 
These prompts (and optional synonyms) are encoded with EVA-CLIP’s text encoder to yield text embeddings
\begin{equation}
    u_c = \text{EVA-CLIP}_\text{txt}(t_c) \in \mathbb{R}^{1024}, \quad c = 1,\ldots,C.
\end{equation}
Region–text similarity in the shared embedding space then provides both a label $c_k$ and a confidence score $s_k$:
\begin{align}
    c_k &= \arg\max_{c} \cos(v_k, u_c), \\
    s_k &= \max_{c} \cos(v_k, u_c),
\end{align}
where $\cos(\cdot,\cdot)$ is cosine similarity. 
Thus, $c_k$ is the class whose description best matches the region, and $s_k$ measures how strongly the region supports that class.

\paragraph{Confidence filtering and pseudo-label bank --}
Because vision–language predictions can be noisy under domain shift, we do not use all region assignments directly. 
Instead, we filter them using a confidence threshold $\tau$:
\begin{equation}
    \mathcal{R}_\text{accept} = \{k \mid s_k \geq \tau\}.
\end{equation}
The accepted regions are then fused into a pixel-wise pseudo-label map $\hat{y}^T \in \{0,1,\ldots,C\}^{H\times W}$, where 0 denotes “unlabeled”. 
In pixels where multiple accepted masks overlap, we break ties by choosing the label from the region with the highest $s_k$:
\begin{equation}
    \hat{y}^T_p = \operatorname*{argmax}_{k: p \in M_k,\; k \in \mathcal{R}_\text{accept}} s_k.
\end{equation}
Pixels not covered by any accepted region remain unlabeled and will be ignored in the target loss. 
The resulting sparse label maps, together with the associated confidences, constitute a \emph{pseudo-label bank} for the target domain.

\paragraph{Optional BLIP refinement of low-confidence regions --}
For regions that fall below the confidence threshold ($s_k < \tau$), we optionally apply BLIP-2 (FLAN-T5 XL)~\cite{li_blip_2023} to obtain a richer, sentence-level description:
\begin{equation}
    g_k = \text{BLIP-2}(x_k^T) = \text{“[caption describing region content]”}.
\end{equation}
We encode the caption with the EVA-CLIP text encoder and compare it against the class prompts:
\begin{equation}
    c_k^\text{BLIP} = \arg\max_c \cos(\text{EVA-CLIP}_\text{txt}(g_k), u_c), \quad
    s_k^\text{BLIP} = \max_c \cos(\cdot).
\end{equation}
A low-confidence region is then reconsidered if the caption-derived label is sufficiently confident or agrees with the original prediction:
\begin{equation}
    s_k^\text{BLIP} \geq \tau_\text{BLIP}
    \quad \text{or} \quad
    (s_k^\text{BLIP} \geq 0.7\,\tau \ \text{and}\ c_k^\text{BLIP} = c_k).
\end{equation}
Such regions are added back into $\mathcal{R}_\text{accept}$ and contribute to the pseudo-label bank. 
This verification-and-fusion step corresponds to the “Pseudo-Label Verification \& Fusion” and BLIP blocks in~\autoref{fig:pipeline}.

\subsection{Stage IV: Pseudo-Label Bank and Downstream Self-Training}
\label{sec:stage_four}

In the final stage, we use the pseudo-label bank to adapt the segmentation model to the target domain. 
We instantiate the segmenter as DeepLabV3~\cite{chen_rethinking_2017} with a ResNet-50 backbone and train it in two phases:

\begin{enumerate}
    \item \textbf{Source pretraining}: supervised training on $D_S$ using cross-entropy and Dice losses, yielding a source-only baseline.
    \item \textbf{Joint self-training}: training on a mixture of labeled source samples and pseudo-labeled target samples. The overall objective is
    \begin{equation}
        L(\theta) = \mathbb{E}_{(x^S,y^S)\sim D_S} [\ell(f_\theta(x^S), y^S)]
                  + \lambda\, \mathbb{E}_{(x^T,\hat{y}^T)\sim \hat{D}_T} [\ell(f_\theta(x^T), \hat{y}^T)],
    \end{equation}
    where the target loss is computed only on pixels that have labels in the pseudo-label bank, and $\lambda$ controls the strength of target supervision.
\end{enumerate}

This stage corresponds to the “Target-Domain Segmentation” block in~\autoref{fig:pipeline}. 
The resulting adapted model is finally evaluated on held-out target validation images and compared against the source-only baseline and alternative adaptation strategies in the evaluation and analysis stage.

\section{Experiments}

We evaluate the proposed cross-modal pseudo-labeling pipeline in two unsupervised domain adaptation settings. 
First, we consider the synthetic-to-real benchmark GTA5-to-Cityscapes, where photo-realistic game-engine imagery serves as the source and real urban scenes as the target. 
The main focus of this paper, however, is a lab-to-factory adaptation scenario for industrial waste sorting, with lab-captured waste images as source and conveyor-belt images from an operational facility as target.
This setting reflects our intended deployment use case and serves as the primary test case for assessing the robustness of the cross-modal pseudo-labeling pipeline under strong real-world domain shift.

\subsection{Datasets and Models}

We present the experimental setup in this section, including datasets, evaluation metrics, and baselines used to assess our proposed cross-modal pseudo-labeling pipeline across different domain shifts, including waste sorting.

\textbf{Datasets --}
We evaluate our proposed pipeline on two domain shifts with increasing practical relevance, chosen to cover both a well-established synthetic-to-real benchmark and a deployment-driven industrial scenario.

\textit{GTA5-to-Cityscapes}~\cite{cordts_cityscapes_2016,richter_playing_2016} (also GTA5$\rightarrow$Cityscapes) is one of the most prominent synthetic-to-real UDA benchmarks for urban scene segmentation. 
\textit{GTA5}~\cite{richter_playing_2016} provides 24{,}966 densely labeled synthetic driving images (18{,}999 train, 5{,}967 val) with 19 urban scene classes. 
\textit{Cityscapes}~\cite{cordts_cityscapes_2016} contains real-world street scenes with fine pixel-wise annotations; the 2{,}975 training images are treated as unlabeled target data for pseudo-label generation, and the 500 validation images are used exclusively for evaluation. 
The domain shift here arises from differences in texture, illumination, and rendering artifacts between the game engine and the real world.


%
%

\textit{LabWaste-to-RealWaste}~\cite{roming_packwise_2025,funk_evaluation_2026} (also LabWaste$\to$RealWaste) is a novel industrial waste sorting benchmark representing lab-to-factory transfer and is the primary focus of this work. 
The source lab waste dataset (\textit{LabWaste})~\cite{roming_packwise_2025} contains approximately 2{,}000 lab-collected images of waste items placed on largely uniform backgrounds, annotated with 8 foreground classes (e.g., plastic, paper, glass, organic) and a background class. 
The target real-world-waste dataset (\textit{RealWaste})~\cite{funk_evaluation_2026} comprises 457 conveyor-belt images from an operational recycling facility, exhibiting motion blur, reflections, dirt, partial occlusions, and complex multi-object scenes. 
This leads to a stronger, more heterogeneous shift than in GTA5-to-Cityscapes and closely reflects our intended deployment conditions.

\textbf{Metrics --} 
Our primary metric is mean Intersection-over-Union (mIoU) across all evaluation classes on the target validation sets. 
As secondary diagnostics, we report per-class IoU to highlight which categories benefit most from adaptation, pseudo-label quality measures (precision, recall, and coverage on labeled target subsets), and the frequency of high-confidence errors, available in our repo.



\textbf{Baselines --}
We compare the cross-modal pseudo-labeling pipeline against several reference methods. 
\textit{Source-only} trains DeepLabV3-R50 on the labeled source domain $D_S$ and evaluates it directly on the target domain $D_T$ without any adaptation. 
%
%
For the GTA5-to-Cityscapes benchmark, we additionally report literature results for \textit{CBST}~\cite{zou_unsupervised_2018}, a class-balanced self-training approach, \textit{DAFormer}~\cite{hoyer_daformer_2022}, a transformer-based UDA method with strong architectural and training enhancements, and \textit{PLSR} (Pseudo Label Self-Refinement)~\cite{zhao_unsupervised_2024}, which refines pseudo-labels via an auxiliary refinement network to better handle label noise. 
These methods rely on larger backbones and task-specific fine-tuning schedules that are carefully optimized for the GTA5-to-Cityscapes setting, so their absolute performance is not directly comparable to our lightweight DeepLabV3-R50 setup; we therefore use them primarily as context for where our approach sits relative to stronger, heavily engineered baselines.

\subsection{Results}
We present the results in three parts: (1) pseudo-label quality as a verifier adaptation and optional BLIP refinement, (2)~target-domain segmentation performance under self-training, and (3)~a focused analysis of the industrial waste dataset. 


\subsubsection{Pseudo-Label Quality Analysis}
\label{sec:plqa}

We evaluate pseudo-label quality on the target validation split to decouple downstream gains from the segmentation optimizer and to identify which configurations are viable teachers for self-training.
We report coverage (fraction of evaluable pixels that receive a pseudo-label after filtering) and per-labeled-pixel correctness (pixel accuracy and mIoU computed only on pseudo-labeled pixels). 
These diagnostics expose the key trade-off in self-training: more labels help only if they are reliable, since structured noise is reinforced across iterations and can lead to confirmation bias~\cite{zhao_unsupervised_2024}.
The standard GTA5$\to$Cityscapes is used as the domain adaptation benchmark; however, we place primary emphasis on LabWaste$\to$RealWaste as the deployment-motivated setting, since our pipeline targets industrial adaptation scenarios in which supervision must be generated automatically from unlabeled target streams.

We report pseudo-label quality under three regimes: no source training (frozen), LoRA-based~\cite{hu_lora_2022} source training that adapts only a few parameters (LoRA), and fully fine-tuned training (Full FT).
We note that verifier adaptation substantially improves region-level recognition on the source domains: on GTA5~\cite{richter_playing_2016}, EVA-CLIP improves Top-1 label accuracy from 0.542 (frozen) to 0.934 (LoRA) and to 0.960 (Full FT); on LabWaste~\cite{roming_packwise_2025}, accuracy rises from 0.325 (frozen) to 0.570 (LoRA) and to 0.600 (Full FT). 
This indicates that stronger adaptation increases the fraction of regions that remain both confident and correct after filtering and fusion, yielding a higher-quality pseudo-label bank for self-training. 
In practice, full fine-tuning provides both higher usable coverage and markedly better per-labeled-pixel correctness, whereas LoRA improves upstream assignments but leaves fewer predictions that pass the confidence as reliable labels.

%

\begin{table}[ht]
\centering
\caption{Pseudo-label quality on target validation: CLIP-only vs.\ CLIP+BLIP. Metrics are computed on pseudo-labeled pixels only. BLIP refinement consistently improves coverage and correctness, with the largest gains under full fine-tuning.}
\label{tab:pl-quality}

\begin{tabular}{@{}lcccccc@{}}
\toprule
\multirow{2}{*}{Method} 
  & \multicolumn{3}{c}{\textbf{GTA5$\to$Cityscapes}} 
  & \multicolumn{3}{c}{\textbf{LabWaste$\to$RealWaste}} \\
\cmidrule(lr){2-4} \cmidrule(lr){5-7}
  & Coverage & Pixel Acc & mIoU & Coverage & Pixel Acc & mIoU \\
\midrule
CLIP (LoRA)          & 4.1\%  & 32.7\% & 3.0\%  & 87.0\% & 4.6\%  & 10.3\% \\
CLIP (Full FT)       & 6.3\%  & 46.3\% & 18.5\% & 92.0\% & 6.3\%  & 16.6\% \\
CLIP+BLIP (LoRA)     & 8.7\%  & 45.9\% & 3.4\%  & 91.5\% & 5.0\%  & 14.0\% \\
CLIP+BLIP (Full FT)  & 10.2\% & 61.6\% & 29.5\% & 96.3\% & 6.5\%  & 21.3\% \\
\bottomrule
\end{tabular}
\end{table}


\autoref{tab:pl-quality} summarizes pseudo-label quality for CLIP-only versus CLIP+BLIP verification.
On GTA5$\to$Cityscapes, accepted pseudo-labels are sparse after filtering (coverage $4.1$--$10.2\%$), reflecting the strict precision requirement under synthetic-to-real shift, where errors on thin objects and boundaries can dominate training.
As an initial parameter-efficient adaptation attempt, LoRA provides slightly higher coverage than the source-only but very low correctness (mIoU $3.0$--$3.4\%$), suggesting that many additional assignments occur in ambiguous regions and are effectively routed to \emph{unknown}/unlabeled outcomes during fusion or removed by confidence filtering.
This motivated moving to full verifier fine-tuning (FT): CLIP (Full FT) substantially improves correctness (mIoU $18.5\%$), and BLIP verification further increases both coverage ($6.3\%\rightarrow 10.2\%$) and correctness (mIoU $18.5\%\rightarrow 29.5\%$, pixel accuracy $46.3\%\rightarrow 61.6\%$), indicating that caption-based grounding can recover usable supervision for otherwise ambiguous regions.

On LabWaste$\to$RealWaste, coverage is high across all settings ($87.0$--$96.3\%$) because the proposal-and-verifier chain labels large parts of the image.
Here, the main challenge is reliability: a substantial fraction of labeled pixels can correspond to belt texture, dirt, shadows, or partial occlusions rather than foreground items.
Moreover, ambiguity is often absorbed by the catch-all class \texttt{other\_packaging}, which increases apparent coverage but does not necessarily improve fine-grained class supervision.
Verifier adaptation improves pseudo-label quality, and BLIP provides consistent gains in labeled mIoU for both full fine-tuning ($16.6\%\rightarrow 21.3\%$) and LoRA ($10.3\%\rightarrow 14.0\%$), consistent with language grounding acting as a selective verifier that resolves some confusions between visually similar packaging categories and rejects a subset of unreliable assignments.

\subsubsection{Target-Domain Segmentation Performance}
\label{sec:tdsp}

\autoref{tab:main} reports target-domain segmentation performance of DeepLabV3-ResNet50 under four regimes: no target adaptation (source-only), self-training with LoRA-based~\cite{hu_lora_2022} pseudo-labels (LoRA), self-training with fully fine-tuned pseudo-labels (Full FT), and self-training with fully fine-tuned pseudo-labels (Full FT) and BLIP verification and expansion for ambiguous labels.

On GTA5-to-Cityscapes, LoRA shows no measurable gain ($20.0\% \to 20.0\%$), whereas Full FT yields a $+25\%$ relative mIoU improvement ($20.0\% \to 25.0\%$).  Adding BLIP verification further improves performance to $26.3\%$ mIoU, corresponding to a $+31.5\%$ relative gain over the source-only baseline. The additional BLIP gain is therefore positive but moderate on Cityscapes.
This asymmetry, where LoRA pseudo-labels improve performance on the waste benchmark but not on Cityscapes, indicates that the Cityscapes configuration is \emph{precision-limited}: structured pseudo-label noise (e.g., boundary leakage, systematic class confusions) can entirely offset the benefits of additional supervision. 
The higher source-only baseline ($20.0\%$ vs.\ $7.7\%$) raises the precision requirements that pseudo-labels must meet to enable positive transfer.
On the industrial waste benchmark (LabWaste-to-RealWaste), LoRA yields a smaller but still meaningful improvement ($+21\%$, $7.7\% \to 9.3\%$), indicating that even imperfect pseudo-labels can provide useful supervision when the source-only baseline is weak. 
Full FT achieves a $+83\%$ relative mIoU improvement ($7.7\% \to 14.1\%$), demonstrating that foundation-model-driven pseudo-labeling can provide substantial gains even when absolute performance remains modest due to the severe operational shift. 
Full FT with BLIP verification achieves the strongest pseudo-label-only result, increasing mIoU to $18.0\%$ and yielding a $+133.8\%$ relative improvement over source-only. This larger gain indicates that caption-based verification is particularly useful in the industrial setting, where visual ambiguity, background clutter, occlusion, and fine-grained packaging categories make contrastive region--text matching alone less reliable.
The low source-only baseline ($7.7\%$ mIoU) leaves substantial room for pseudo-label-driven adaptation, which explains why both verifier variants improve.

%
\begin{table}[t]
\centering
\caption{Target-domain segmentation without and with AdaBN initialization. AdaBN alone substantially improves over source-only by reducing representational mismatch. Combining AdaBN with high-quality pseudo-labels (Full FT) yields additive gains; LoRA pseudo-labels provide smaller improvements.} 
\label{tab:main}
\begin{tabularx}{\textwidth}{@{}lXXXX@{}}
\toprule
\multirow{2}{*}{Method} 
  & \multicolumn{2}{c}{\textbf{GTA5$\to$Cityscapes}} 
  & \multicolumn{2}{c}{\textbf{LabWaste$\to$RealWaste}} \\
\cmidrule(lr){2-3} \cmidrule(lr){4-5}
  & mIoU & Rel.\ gain & mIoU & Rel.\ gain \\
\midrule
Source-only    & 20.0 & -- & 7.7 & -- \\
LoRA        & 20.0 & +0\%   & 9.3  & +21\%  \\
Full FT & 25.0 & +25\%  & 14.1 & +83\%  \\
Full FT + BLIP verification & 26.3 & +31.5\% & 18.0 & +133.8\% \\
AdaBN + Source-only & 33.0 & +65.0\% & 27.4 & +255.8\% \\
AdaBN + LoRA & 34.1 & +70.5\% & 28.0 & +263.6\% \\
AdaBN + Full FT & \textbf{36.0} & \textbf{+80.0\%} & \textbf{30.2} & \textbf{+292.2\%} \\
\midrule
CBST~\cite{zou_unsupervised_2018}      & 45.2 & -- & -- & -- \\
DAFormer~\cite{hoyer_daformer_2022}    & 68.3 & -- & -- & -- \\
PLSR~\cite{zhao_unsupervised_2024}    & 71.3 & -- & -- & -- \\
\bottomrule
\end{tabularx}
\end{table}
\paragraph{Isolating domain alignment from pseudo-label effects}
To disentangle gains attributable to pseudo-label supervision from gains due to implicit domain alignment (e.g., exposure to target statistics), an AdaBN~\cite{li_adaptive_2018} control is evaluated in which only BatchNorm running statistics are updated on unlabeled target images, without any semantic supervision. 
It should be noted that BLIP verification is evaluated in the pseudo-label self-training branch without AdaBN initialization. The AdaBN rows isolate the effect of target-domain normalization statistics, whereas the BLIP row isolates the effect of language-based pseudo-label verification. 
Results are reported in~\autoref{tab:main}.

On Cityscapes, AdaBN increases mIoU from $20.0\%$ to $33.0\%$; adding LoRA on top of AdaBN yields $34.1\%$, whereas Full FT reaches $36.0\%$. 
The relative ordering of pseudo-label variants is preserved under both initializations, confirming that pseudo-label reliability remains the main limitation regardless of initialization quality.
On the industrial waste benchmark, AdaBN alone substantially improves target performance ($7.7\% \to 27.4\%$ mIoU), indicating that a large part of the source–target gap arises from representational mismatch in the segmentation backbone rather than from missing semantic supervision. 
Importantly, when AdaBN is combined with high-quality pseudo-labels, additive improvements are observed: AdaBN + LoRA reaches $28.0\%$ mIoU, while AdaBN + Full FT reaches $30.2\%$. 
These results show that pseudo-label reliability remains a limiting factor even with a stronger, target-aligned initialization, and that high-quality pseudo-labels provide benefits beyond what domain alignment alone can achieve.



\begin{figure}[t]
    \centering
    \begin{subfigure}[b]{0.24\textwidth}
        \includegraphics[width=\textwidth,clip,trim=0 0 1900 0]{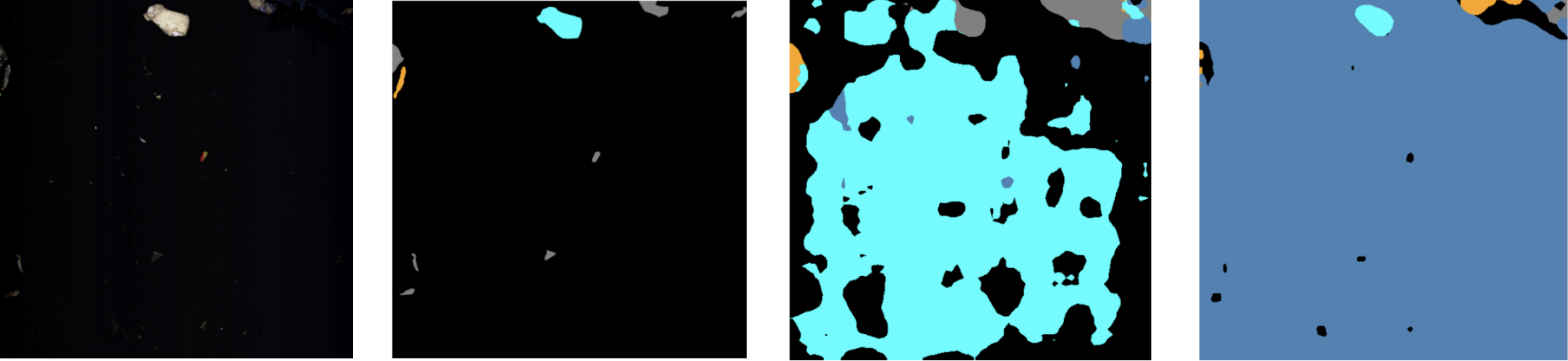}
        \caption{Raw image}
        \label{fig:waste-a}
    \end{subfigure}
    \hfill
    \begin{subfigure}[b]{0.24\textwidth}
        \includegraphics[width=\textwidth,clip,trim=620 0 1280 0]{figures/waste-masks.png}
        \caption{GT mask}
        \label{fig:waste-b}
    \end{subfigure}
    \hfill
    \begin{subfigure}[b]{0.24\textwidth}
        \includegraphics[width=\textwidth,clip,trim=1240 0 650 0]{figures/waste-masks.png}
        \caption{No pseudo-labels}
        \label{fig:waste-c}
    \end{subfigure}
    \hfill
    \begin{subfigure}[b]{0.24\textwidth}
        \includegraphics[width=\textwidth,clip,trim=1890 0 0 0]{figures/waste-masks.png}
        \caption{Pseudo-labels}
        \label{fig:waste-d}
    \end{subfigure}
    \caption{Qualitative example on the industrial waste benchmark. (a) Raw conveyor-belt image, (b) ground-truth semantic mask, (c) segmentation output of the DeepLabV3-R50 model, and (d) segmentation output of the adapted DeepLabV3-R50 model using our proposed pipeline (more similar to the GT mask)}
    \label{fig:waste-masks-qual}
    \vspace{-1em}
\end{figure}

\subsubsection{Industrial Deployment: Plastic Waste Sorting --}
\label{sec:case}


The LabWaste-to-RealWaste~\cite{roming_packwise_2025,funk_evaluation_2026} benchmark represents a practically relevant lab-to-factory transfer scenario in industrial plastic waste sorting, where a model trained on controlled laboratory data must generalize to operational conditions. 
Several characteristics differentiate this scenario’s setting from a synthetic to a real-world problem, creating unique challenges for pseudo-label-driven adaptation.

Most notably, conveyor-belt imagery, as in our waste sorting datasets~\cite{roming_packwise_2025,funk_evaluation_2026}, is dominated by background pixels; foreground objects are sparse and often small (\autoref{fig:waste-a}), which inflates coverage metrics (\autoref{tab:pl-quality}) while making per-class correctness critical for practical utility. 
Unlike urban driving scenes, where multiple object categories co-occur densely, the waste setting requires the pipeline to avoid false positives on large homogeneous regions while still detecting the relatively rare foreground items (\autoref{fig:waste-b}). 
With background pixels dominating, aggregate pixel accuracy tends to represent a clean laboratory source domain, leaving the LabWaste-to-RealWaste shift both challenging and lacking real-world value.
Compounding this challenge, target images exhibit substantial visual contamination: dirt, shadows, specular reflections, and variations in belt texture create spurious edges and local appearance patterns that can trigger false mask proposals or lead the verifier to assign confident but incorrect labels to background regions.
These artifacts are absent from the clean laboratory source domain, making the LabWaste-to-RealWaste shift challenging for both the SAM proposal and EVA-CLIP verification stages. 

Also, the waste classes include fine-grained packaging categories (e.g., cup/tray vs.\ lid/cap, bottle vs.\ can) that are visually similar under partial occlusion and contamination. 
This class ambiguity leads to systematic confusions that can propagate through self-training if pseudo-labels are not sufficiently filtered. 
The catch-all category (\textit{other\_packaging}) collects many predictions, which explains its disproportionately high per-class IoU relative to fine-grained categories.

Despite these challenges, the pipeline achieves its largest relative gains on the industrial waste benchmark: Full FT improves mIoU from $7.7\%$ to $14.1\%$ (+83\% relative), exceeding the +25\% relative gain observed on Cityscapes. 
This asymmetry is informative: when the source-only baseline is weak and the target domain is severely underspecified, even imperfect pseudo-labels provide useful supervision, whereas on Cityscapes, the higher baseline and stricter precision requirements leave less room for noisy labels to help. 
Qualitative inspection shows that self-training suppresses large-scale false positives on conveyor-belt texture while preserving salient foreground objects (\autoref{fig:waste-c} vs. \autoref{fig:waste-d}), with remaining errors dominated by missed small fragments and under-segmentation of rare classes, consistent with the conservative confidence filtering needed to maintain pseudo-label reliability in this high-ambiguity setting.

From a deployment perspective, these results suggest that foundation-model-driven pseudo-labeling can reduce annotation effort for industrial applications. 
Instead of collecting dense pixel-wise labels for every new factory site, a process that demands domain expertise, specialized tooling, and repeated updates as conditions change, practitioners can use pseudo-label self-training to obtain a stronger initialization than source-only transfer. 
While absolute performance remains modest and further domain-specific refinement may be required for safety-critical use, the method offers a practical path toward automatic annotation in scenarios where manual labeling is infeasible at scale.

\section{Discussion}

Across both benchmarks, the main gains come from an end-to-end pseudo-labeling pipeline that derives target-domain supervision from unlabeled images via foundation models, rather than from the segmenter’s own predictions. 
SAM supplies geometry-driven region proposals, EVA-CLIP assigns semantics via region–text similarity, and the resulting pseudo-label bank enables self-training with an external supervision signal that is less tied to the segmenter’s often miscalibrated confidence under shift. 
Empirically, configurations that produce more reliable pseudo-labels yield larger target-domain gains, whereas increasing pseudo-label amount without improving accuracy offers limited benefit. 
We use a standard DeepLabV3-R50 backbone to keep the pipeline deployable and to isolate the effect of the pseudo-labeling chain instead of architectural advances.

A key observation is the mismatch between verifier-level accuracy and segmentation gains. 
LoRA can approach full fine-tuning on region classification yet still yield weaker downstream improvements under stronger shifts, reflecting self-training’s sensitivity to structured pseudo-label noise. 
Spatially correlated boundary leakage and systematic confusions (e.g., thin objects, vehicle subclasses, or visually similar packaging categories) are reinforced during optimization, so a verifier that is “good on average” can still be a poor teacher if its residual errors are class-biased or concentrated on boundaries and small objects.

The industrial waste setting illustrates the applied motivation: shifts arise from changing acquisition conditions (lighting, belt texture, and contamination), occlusion, and site-specific artifacts, while object appearance evolves over time. 
In such regimes, dense segmentation labels are costly, and our results suggest that a foundation-model-driven pipeline can generate supervision from unlabeled target streams with minimal manual effort, provided pseudo-label reliability is controlled through adaptation and filtering. 
The design exposes practical knobs, such as proposal pruning and confidence thresholds, that can be tuned per site without altering the training objective or requiring target annotations.

Two implementation-level constraints directly influence pseudo-label usefulness and compute cost: 
\textit{(i) proposal density control}, since automatic SAM proposals can be highly redundant and require a cap (top-$K$ after filtering) plus overlap suppression to avoid noisy, inefficient supervision; 
and \textit{(ii) overconfidence under shift}, as EVA-CLIP similarities can remain high for incorrect assignments, making a single global threshold insufficient and motivating class-based thresholds, or simple consistency checks to reduce confidently wrong labels.

\textbf{Limitations and threats to validity --}
The BLIP extension is most effective as a selective verifier for ambiguous regions rather than a general relabeler. 
It can improve pseudo-label performance when the vision–language assignment is uncertain, but it incurs computational cost, and its benefit depends on careful triggering (low-confidence crops only) and strict acceptance criteria. 
In this work, we evaluate BLIP at the pseudo-label level as an optional verification module and examine whether full end-to-end retraining with BLIP-verified labels also improves target performance.
Our approach assumes a closed-label set at training time; open-set items are currently routed to an `unknown' bucket rather than used to expand the label set. Secondly, performance remains limited on thin structures and small objects, where proposal quality and label assignment are most fragile.
Thirdly, while we compare our methods to a baseline DeepLabV3, comparisons to UDA state-of-the-art should be interpreted with caution: methods like DAFormer use stronger backbones and specialized training strategies, whereas our goal is to quantify how far an off-the-shelf segmenter can be pushed with foundation-model supervision in an end-to-end, deployment-oriented pipeline.

\textbf{Future work --} 
Building on the modularity of our pipeline, a next step is to systematically evaluate stronger and more diverse segmentation backbones, such as SegFormer~\cite{xie_segformer_2021} or ViT-based architectures, as drop‑in replacements for DeepLabV3‑R50 to better disentangle backbone capacity from the benefits of cross‑modal pseudo‑labels and to narrow the gap to transformer‑based UDA methods like DAFormer~\cite{hoyer_daformer_2022} and PLSR~\cite{zhao_unsupervised_2024}. 
Beyond external supervision alone, future work could combine our vision–language pseudo‑labels with internal refinement strategies (e.g., self‑training from the segmenter’s own predictions or open‑vocabulary segmentation backbones), enabling hybrid approaches that adaptively balance external CLIP pseudo-labels and model confidence under different shift regimes. 
Finally, to strengthen claims about robustness and practical deployment, we plan to extend the industrial evaluation to larger, multi‑site, and temporal splits, and to fine‑tune representative UDA baselines and open‑vocabulary segmenters on these different domains.

\section{Conclusion}

We presented a pseudo-labeling pipeline for unsupervised domain adaptation in semantic segmentation that requires no target-domain labels and serves as a robust, modular alternative that practitioners can easily integrate into existing workflows. 
Foundation models are used throughout: SAM generates class-agnostic masks, and EVA-CLIP assigns semantic labels via vision–language matching, while remaining interchangeable with stronger models in the future. 
A confidence filter balances coverage and reliability, and an optional BLIP-based verification stage refines ambiguous regions without changing the self-training loop.

The pipeline is evaluated on synthetic-to-real driving scenes (GTA5-to-Cityscapes) and on lab-to-factory industrial waste sorting (LabWaste-to-RealWaste), where it outperforms source-only training. 
In real-world settings, we find that pseudo-label quality, rather than quantity, governs self-training success under domain shift: verifiers with similar accuracy can have very different downstream effects when their errors are structured and reinforced during optimization. 
This suggests that practical effort is best invested in improving pseudo-label fidelity, via verifier adaptation, cross-modal checks, or calibrated filtering, so that self-training can genuinely reduce annotation effort, especially in underspecified industrial target domains where baselines are weakest.

%
%

\begin{credits}
\subsubsection{\ackname} 
We thank the Fraunhofer Institute of Optronics, System Technologies and Image Exploitation (IOSB) and Lobbe RSW GmbH for providing the lab- and real-world lightweight packaging waste dataset.  
The authors used AI-assisted tools to improve the language and readability of this paper. 
All content was reviewed, edited, and approved by the authors, who take full responsibility for the final text.
\end{credits}

%
%
%
\bibliographystyle{splncs04}
\bibliography{small-ecml2026}
%




\end{document}